\documentclass{article}
\usepackage{graphicx}
\usepackage{authblk}

\usepackage{placeins}
\usepackage{hyperref}
\usepackage{float}
\usepackage{amssymb,amsthm,amsmath}
\usepackage{hyperref,titlesec}
\usepackage{cleveref}
\usepackage{sectsty}

\usepackage[authoryear]{natbib}
\setcitestyle{braces=round}
\usepackage{subcaption}

\titleformat{\section}[display]{\normalfont\huge\bfseries\centering}{\centering\chaptertitlename\thechapter}{10pt}{\Large}
\titlespacing*{\section}{0pt}{0ex}{0ex}

\newcommand\citex[1]{\textit{\citeauthor{#1}}}

\begin{document}
\title{The Cognate Data Bottleneck in Language Phylogenetics} 

\author[1]{Luise H{\"a}user}
\author[2]{Alexandros Stamatakis}

\affil[1]{Computational Molecular Evolution group, \\
  Heidelberg Institute for Theoretical\\
  Studies, Heidelberg, Germany, \\
  Institute for Theoretical Informatics, \\
  Karlsruhe Institute of Technology, Karlsruhe, Germany \\
  \texttt{luise.haeuser@h-its.org}}


\affil[2]{Biodiversity Computing Group, \\
  Institute of Computer Science, Foundation for Research and Technology - Hellas\\
  Computational Molecular Evolution group, \\
  Heidelberg Institute for Theoretical Studies, Heidelberg, Germany
  Institute for Theoretical Informatics, \\
  Karlsruhe Institute of Technology, Karlsruhe, Germany \\
  \texttt{stamatak@ics.forth.gr}}

\date{\today}
\maketitle

\begin{abstract}
To fully exploit the potential of computational phylogenetic methods for cognate data one needs to leverage specific (complex) models an machine learning-based techniques. However, both approaches require datasets that are substantially larger than the manually collected cognate data currently available. To the best of our knowledge, there exists no feasible approach to automatically generate larger cognate datasets. We substantiate this claim by automatically extracting datasets from BabelNet, a large multilingual encyclopedic dictionary. We demonstrate that phylogenetic inferences on the respective character matrices yield trees that are largely inconsistent with the established gold standard ground truth trees. We also discuss why we consider it as being unlikely to be able to extract more suitable character matrices from other multilingual resources. Phylogenetic data analysis approaches that require larger datasets can therefore not be applied to cognate data. Thus, it remains an open question how, and if these computational approaches can be applied in historical linguistics. 
\end{abstract} 

\bigskip

\noindent

\section{Introduction}

Originally developed for analyzing biological data, computational phylogenetic methods are now also routinely being used in historical linguistics. In this field, phylogenetic methods such as Bayesian Inference \cite{kolipakam18, sagart19, heggarty23} or Maximum Likelihood based tree inference \cite{jaeger18} are predominantly applied to cognate data \cite{dunn13}. The cognate datasets typically encoded as binary character matrices \cite{haeuser24} to be provided as input to the inference tools. 
While this binary encoding is straightforward, it also has its drawbacks\cite{haeuser25-3, evans06}. If cognate data are encoded in a more sophisticated (i.e., in a non-binary) manner, the analyses require a distinct evolutionary model \cite{haeuser25-3}. However, such a model will comprise more free parameters that in turn will require a larger amount of cognate data to be reliably estimated in order to circumvent overparametrization. In addition, recent advances in phylogenetics increasingly rely on machine learning techniques \cite{haag22, trost23, azouri21, nesterenko24}. Applying these approaches to cognate data also necessitates large datasets to obtain accurate and robust results. The currently available cognate datasets are manually assembled and are hence substantially smaller than molecular datasets \cite{haeuser25-3}.
This raises the question, whether one can create substantially larger cognate datasets. Here, we investigate if this can be achieved in a fully automated manner. To this end, we extract cognate character matrices from the multilingual encyclopedic dictionary BabelNet \cite{navigli12}. The extraction process poses several challenges that we cannot overcome without introducing errors and thereby, decreasing data quality. Whether we can retrieve a sufficient amount of data to compensate for this, constitutes a fundamental question we will address.
The remainder of this paper is organized as follows: First, we discuss potential sources and approaches for automatically obtaining additional, large cognate data in \cref{sec:sources}. We provide a rationale why we chose BabelNet for data extraction. Subsequently, we present BabelNet in greater detail in \cref{sec:babelnet} and explain how we extract data to obtain character matrices (\cref{sec:challenges}). We assess these character matrices by analyzing their completeness (\cref{sec:compeleteness}) and by inferring Maximum Likelihood (ML) trees which we compare to the Glottolog gold standard trees (\cref{sec:signal}). Unfortunately, our experiments reveal an insufficient quality of the automatically extracted BabelNet data. We also assess the errors that the individual components of this automatic data generation pipeline introduce (see \cref{sec:reverse}). 
Given the aforementioned possible data sources, it is likely that analogous challenges will occur when extracting character matrices from a different linguistic resource. Therefore, we have currently reached a limit regarding the application of more sophisticated phylogenetic inference approaches and models to cognate data. The available datasets are neither sufficient in size and number for applying more complex models or machine-learning based approaches, nor is it possible to automatically acquire the data necessary for such endeavors. We discuss this in detail in (\cref{sec:conclusion}).

\section{Potential Data Sources and Data Selection Strategies}
\label{sec:sources}
In this section, we first outline why it is difficult to identify data sources that are suitable for extracting data and assembling them into datasets that are apt for phylogenetic inference. Then, we discuss some (more or less well-known) multilingual language resources. We assess the respective obstacles to character matrix extraction and construction and justify why we henceforth focus on BabelNet.
\subsection{Challenges}
There exist numerous multilingual resources that serve different purposes. The number and variety of languages covered is pivotal when choosing a resource. Extensive resources are available for at most $100$ of the $7000$ languages being spoken worldwide. The remaining $6900$ are so called low-resource languages \cite{cieri16} for which data availability is suboptimal. As a consequence, these languages are excluded in numerous recently developed Natural Language Processing tools \cite{googhari23}.\\
To be suitable for phylogenetic inference, the data must further fulfill two key requirements. Firstly, they must exhibit \textit{parallelism} between the languages. That is, the data must be structured in a way such that certain features or characteristics can be determined, which can subsequently be mapped to the columns of a character matrix. Secondly, the data provided for a particular feature must be \textit{comparable} among the languages considered that is, a binary or multi-valued encoding must be possible. For cognate data, the features are the concepts that are specified in the concept list of a specific dataset \cite{dunn13}. Hence, a key challenge is to determine an appropriate concept list, as it is hard to identify - even a few - concepts, which are universal to all languages \cite{evans06}. Once such a concept list has been established, comparability can be attained by determining the cognate classes of the words and by creating the corresponding binary presence-absence matrices \cite{haeuser24}. Note that, parallel words provided in the orthography of the respective language do not ensure comparability. The English word \textit{bacterium} is for example related to the Greek word $\beta\alpha\kappa\tau\eta\rho\iota o$. The similarity of the words is not revealed via direct comparison as the languages use different scripts. Therefore, additional phonetic information is required to determine cognate classes or to compare data from different languages in any meaningful manner \cite{jaeger18}. As mentioned above, it remains an open question which alternative possibilities exist for aligning lexical data. There exist approaches relying on alignments of the IPA transcriptions' sound classes \cite{jaeger18,akavarapu24-likelihood} or on the analysis patterns in sound changes \cite{haeuser2024-cp}. Another possible solution is to develop an encoding where each concept is represented in the character matrix by one single column only. However, as outlined in \citex{haeuser25-3}, these alternative representations require more complex models with a larger number of free parameters than the models for binary data. Larger automatically extracted data sets could be used to obtain meaningful estimates of these parameters. Henceforth, we nonetheless restrain ourselves to the standard binary representation, as our focus is on data acquisition.
\subsection{Possible Sources}
Corpora based on parallel texts constitute one potential multilingual data source. The best known one is probably the \textit{Parallel Bible Corpus} \cite{mayer14}. It is based on $900$ Bible translations in $830$ language varieties. The texts themselves are not publicly accessible though due to copyright restrictions. However, two files are publicly available for each Bible version: a wordlist with all words occurring in the text as well as a matrix indicating the number of occurrences for each Bible verse and each word form. The structure of the verses ensures parallelism between the versions. However, we consider it as being too challenging to construct an analogous encoding based on word form occurrences and therefore refrain from further investigating the Parallel Bible Corpus. In order to ensure the parallelism of the data, the word forms would have to be aligned on the basis of the occurrences, which are probably indicated too coarse-grained for this. \\
The corpus \textit{FLORES} \cite{nllb22} comprises $3001$ sentences in $205$ languages retrieved from $842$ manually translated web articles. In comparison to the parallel Bible Corpus, the parallel sentences {\em are} publicly available. However, the transfer of these sentences into a comparable encoding would require extracting parallel words from them. This is not possible without - at least - the availability of part-of-speech-tags, which are however missing in FLORES. The same restriction applies to other parallel text corpora.\\
\newpage
Another group of multilingual resources are benchmark datasets for evaluating Large Language Models (LLMs). They are used to quantify LLM performance on specific tasks where the input is provided in different languages. The \textit{TAXI1500} \cite{ma24} benchmark has been assembled for text classification. It is based on the Parallel Bible Corpus. In TAXI1500, each verse of the English version is annotated by one out of five tags. The annotations are then transferred to the parallel Bible verses in other languages. The benchmark is hence ideal for testing the ability of an LLM to classify sentences in different languages. Nonetheless, it does not contain any information that can be deployed for phylogenetic inference. The same limitation is inherent to \textit{SIB-200} \cite{adelani24} and Belebele \cite{bandarkar23} which are based on the FLORES corpus, and to other resources such as \textit{WikiANN} \cite{pan17}, \textit{MASSIVE} \cite{fitzgerald23}, and \textit{XTREME} \cite{hu20}. All of these resources are multilingual, but they have been specifically developed for benchmarking LLMs and do therefore not contain phylogenetic signal. \\

Numerous tasks in natural language processing, such as natural language understanding \cite{li18}, can be better addressed if one does not directly operate on words, but instead, on word embeddings. Word embeddings are mappings of the words into a vector space that represent their mutual relationships via spatial proximity \cite{almeida23}. To potentially perform phylogenetic inference on word embeddings one requires resources that provide word embeddings in as many languages as possible. \textit{BPEmb} \cite{heinzerling18} contains, for example, word embeddings for words from Wikipedia in $275$ languages. \\
Another option is to utilize multilingual LLMs that have been pre-trained on data in multiple languages. For example, there is a multilingual version of the well-known LLM \textit{BERT} \cite{devlin19}, which relies on training data from over $100$ languages. \textit{XLM-V} \cite{liang23} has been pretrained on numerous different multilingual resources, including FLORES with data from more than $200$ languages. \textit{Glot500} \cite{googhari23} focuses on low-resource languages, covering more than $500$ of them. The training data for \textit{SERENGETI} \cite{adebara23} originates from more than $500$ African languages, many of which are considered as low-resource languages. \\
Applying clustering methods to word embeddings returns sets of words that describe similar concepts \cite{zhang17}. This allows to obtain parallel data for different languages from these word embeddings. However, we do require additional phonetic information to create aligned data from parallel words. Phonetic word embeddings \cite{sharma21,zouhar24} address this problem. However, the field is still in its infancy and phonetic word embeddings are currently only available for $9$ distinct languages.\\

The last group of multilingual resources we consider, comprises semantic networks, knowledge-bases, and encyclopedic dictionaries. Their key advantage is that they are well-structured. The basis for many resources is \textit{WordNet} \cite{miller95}, a semantic network of the English language. \textit{EuroWordNet} \cite{vossen98} represents the first attempt to extend WordNet to distinct languages and to also connect the vocabulary of these languages. However, it only contains data for $7$ languages. \textit{Open Multilingual Wordnet} \cite{bond16} combines different WordNets and supports over $150$ languages. We do not consider this resource further as it lacks phonetic information.\\
Different resources maintained by the \textit{Wikimedia Foundation} can serve as structured multilingual data sources. \textit{Wikipedia} can be used to acquire information about semantic relatedness among words, as shown by \citex{strube06} for the English version. \textit{DBPedia} \cite{lehmann14} offers a semantic network extracted from Wikipedia. However, while there exist Wikipedia versions in $325$ languages (\url{https://de.wikipedia.org/wiki/Wikipedia:Sprachen}), DBPedia only covers $6$ languages \cite{kontokostas12}.\\
\textit{Wiktionary} is a large multilingual dictionary with the goal to provide definitions for all words in all languages. In contrast to most other resources, it has the advantage that phonetic information is (partly) provided in the form of International Phonetic Alphabet (IPA) transcriptions. The English Wiktionary has more than $8$ Million entries in $4400$ languages (\url{https://en.wiktionary.org/wiki/Wiktionary:Main_Page}), additionally, there exist versions in numerous other languages. \textit{DBnary} \cite{serasset12} is based on $22$ of them and provides access to multilingual lexical data, yet for $25$ languages only.\\
\textit{Wikidata} is the last resource of the Wikimedia Foundation we discuss here. It is a large knowledge base that is provided as a graph where each node corresponds to an entity and where edges represent distinct relationships between these entities \cite{suchanek24}. \textit{Yago} \cite{suchanek24} is based on Wikidata and combines it with \textit{Schema.org} \cite{guha15}, a collaboratively developed ontology. It predominantly focuses on providing taxonomically structured language data for question answering or knowledge injection, but not on multilinguality. \\

\textit{ConceptNet} \cite{speer17} is based on data from the Open Mind Common Sense project (\url{https://www.media.mit.edu/projects/open-mind-common-sense/overview/}) and combines it with DBPedia, Wikitionary, Open Multilingual WordNet as well as with a high-level ontology from OpenCyc (\url{https://github.com/asanchez75/opencyc?tab=readme-ov-file}) and also with data collected via a word game for building a large semantic network. In total, it supports $304$ different languages. Despite being a valuable combination of different resources we do not consider using ConceptNet, as it also lacks phonetic information and because it is, unfortunately, no longer supported.\\

\textit{BabelNet} \cite{navigli12, navigli21} is a multilingual encyclopedic dictionary that combines the structure of WordNet with Wikipedia, Wiktionary, Wikidata, and numerous other resources. It contains data for more than $600$ languages. They are structured as a semantic network that has been extended for multilingual purposes. This facilitates obtaining parallel data for different languages. Moreover, phonetic information is at least partially available in the form of IPA transcriptions, so that parallel data can also be aligned. Thus, BabelNet is the only among the data sources we considered that offers both, a structure that is apt for our purposes, and phonetic information. Finally, it also covers a large variety of distinct languages. Therefore, we investigate in more detail, how one can extract data for phylogenetic inference from BabelNet. \\
\newpage
\section{BabelNet}
\label{sec:babelnet}
BabelNet adopts the notion of \textit{synsets} from WordNet to structure its vocabulary. A synset unites words, called \textit{senses} that describe the same concept. In contrast to WordNet, in BabelNet the senses of one synset can originate from different languages \cite{navigli21}. To construct character matrices, we use the parallelism provided by the synset structure. Each synset is thus represented by a group of binary columns in the resulting character matrix.\\
All steps of the data extraction process and the associated challenges are discussed in detail in \cref{sec:challenges} below. To quantify the induced error by automatic IPA transcription and tokenization (see \cref{sec:transcription}), we deploy a reverse engineering approach which we present in \cref{sec:reverse}. In the subsequent \cref{sec:evaluation}, we present the results of different experiments to assess the data quality of these character matrices and their suitability for phylogenetic inference. All described experiments are available on Github (\url{https://github.com/luisevonderwiese/babel2msa/tree/master}). The resulting datasets contain processed data from BabelNet v 5.0 downloaded from \url{https://babelnet.org} and are made available under the BabelNet license (see \url{https://babelnet.org/full-license}).

\subsection{Character Matrix Construction}
\label{sec:challenges}
In this section, we discuss in detail, how we extract character matrices for phylogenetic inference from BabelNet, version 5.3. In \cref{sec:languages}, we describe how we identify and select languages for the resulting datasets. Then, we explain how we choose the synsets that shall be included in the character matrices (see \cref{sec:synset-selection}). As the proportion of available IPA transcriptions is prohibitively small, we automatically transcribe senses into IPA using the epitran \cite{mortensen18} tool. Further, these IPA transcriptions must be automatically tokenized. We explain the details of these steps in \cref{sec:transcription}. In \cref{sec:cognate-clustering} we outline how we automatically cluster cognates for obtaining the character matrices for which we subsequently infer phylogenies via Maximum Likelihood. 

\subsubsection{Selection of Languages}
\label{sec:languages}
In BabelNet, languages are identified by ISO codes, while Glottocodes are required when using Glottolog to determine the language families or to conduct comparisons with the gold standard tree. Mapping the codes is challenging as there only exists an incomplete many-to-many relationship between the two naming systems (\url{(https://clld.org/2015/11/13/glottocode-to-isocode.html)}. 
Henceforth, we only consider the ISO code languages that we can map to a glottocode. \\
We further conduct experiments using data for two language subsets. The first subset contains $161$ Indo-European Languages; this selection is the same as used in the \textit{Indo-European Cognate Relationships database} (\textit{iecor}) \cite{heggarty23}. With the second subset, denoted by \textit{dense} languages in the following, we aim to create a dataset that is as dense as possible. The \textit{dense} dataset includes all languages for which we can automatically transcribe orthographic words into IPA via epitran. The tool supports a total of $94$ languages, of which we use $77$. These $77$ are the languages that are included in BabelNet and have an ISO code that can be mapped to a Glottocode. We refer to these languages as dense because we assume that the resulting character matrices will be densely populated due to epitran support.

\subsubsection{Synset Selection}
\label{sec:synset-selection}
\begin{figure}[h!]
    \centering
      \begin{subfigure}{0.49\textwidth}
        \includegraphics[width=\textwidth]{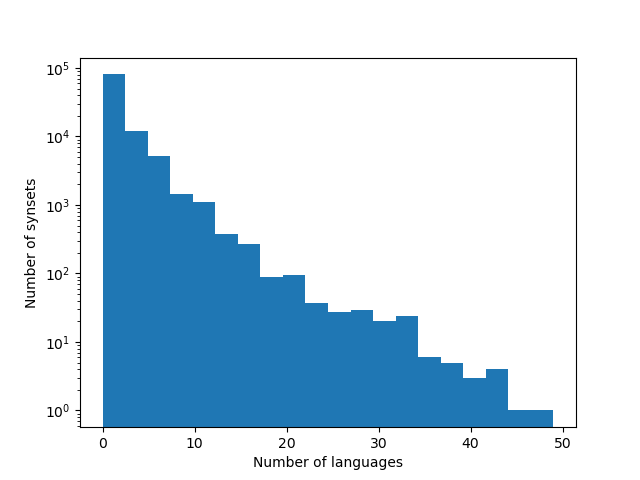}
          \caption{BabelNet}
          \label{fig:filter}
      \end{subfigure}
      \begin{subfigure}{0.49\textwidth}
        \includegraphics[width=\textwidth]{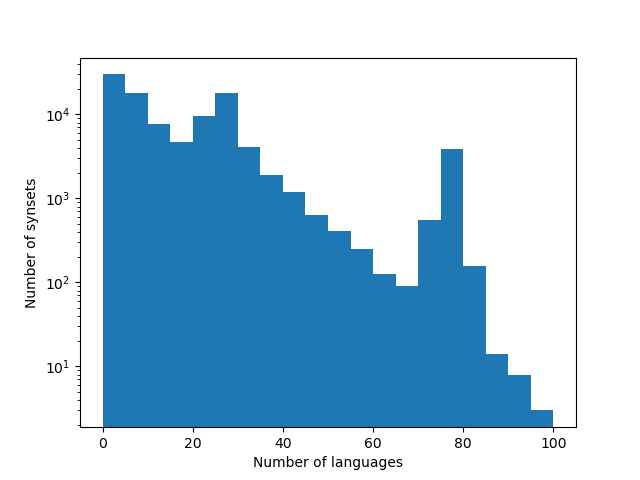}
          \caption{BabelNet + epitran}
          \label{fig:filter-epitran}
      \end{subfigure}
\label{fig:count-hists}
\caption{Data availability in different languages for synsets in BabelNet\\
The number of languages is depicted along the x-axis, the y-axis (using a logarithmic scale) corresponds to the number of synsets, for which there is a main sense available in the respective number of languages. In Subfigure (a) we only count main senses with an IPA transcription provided in BabelNet, in Subfigure (b) we additionally count main senses without an IPA transcription in BabelNet, but with the respective language support in epitran.}
\end{figure}

All synsets in BabelNet are labeled as either \textit{entity} or \textit{concept}. We only use the latter. Named entities are typically described by the same or similar words in different languages. These words all belong to the same cognate class. Subsequently, they are represented by a single column in the corresponding character matrix which contains the value $\mathtt{1}$ for all languages and does hence not contain any phylogenetic signal. Named entities are therefore omitted from the character matrices.\\
For each synset, one specific sense in each language is tagged as \textit{main sense}. This main sense detection is based on an algorithm that considers various factors such as the source the sense has been extracted from, the relevance of the lexicalization in the synset, or the node degree in the semantic network \cite{cecconi24-private}. We exclude senses other than the main sense, as these may comprise highly specific or exotic terms whose use is likely to blur the phylogenetic signal. This approach is analogous to the "most frequent sense heuristic" that is used in word sense disambiguation \cite{raganato17}. As a consequence, the resulting datasets do not contain any synonyms and the character matrices therefore exhibit no polymorphisms. \\
For each synset, we determine the number of languages for which a main sense with an IPA transcription is present. When we use epitran to transcribe orthographic words to IPA (see \cref{sec:transcription}), it suffices to only have one main sense available, as long as the specific language is supported by epitran. Therefore, for each synset, we also count the number of languages, for which we can obtain a main sense {\em with} an IPA transcription either by directly retrieving it, \textit{or} via automatic IPA transcription. However, we still filter out synsets without a main sense having at least one IPA transcription provided in BabelNet in any of the respective languages.\\
\cref{fig:count-hists} illustrates the data availability for BabelNet synsets. In both figures, the number of languages is depicted along the x-axis, while the y-axis corresponds to the number of synsets, for which there exists main sense in the respective number of languages. Note that we use a logarithmic scale for the y-axis. In \cref{fig:filter}, we only count the main senses for which there is an IPA transcription available in BabelNet. For $99.8\%$ of the  synsets, IPA transcriptions are available in only $20$ languages or less. Even the largest synset only covers less than $50$ languages. In \cref{fig:filter-epitran}, we also take into account the main senses without IPA transcription, as long as the respective language is supported by epitran. We observe a shift in the distribution due to the use of automatic IPA transcription. In this case, $2.0\%$ of the synsets become available in more than $60$ languages. However, $60.6\%$ of the synsets still remain small, covering $20$ languages or less. This is mainly due to the lack of phonetic information. On the one hand, there exist only $77$ languages, for which we can obtain an epitran instance {\em and} map the ISO code to a glottocode. Therefore, epitran can only alleviate the data sparsity issue for a small proportion of languages. On the other hand, even when supported by epitran, we still require a main sense to be specified. This is often not the case for many languages. Hence, the number of languages for which we can construct a reasonable character matrix is substantially smaller than the total of $600$ languages available in BabelNet. 
\newpage
Due to the sparsity of the data provided, it is necessary to determine a subset of synsets that maximize the data that are available for the languages under study in order to obtain a character matrix that is as densely populated as possible. \\
To select synsets, we explore two distinct approaches. 
In the first one, we sort the synsets by decreasing sense availability, based on the above counts. Note that the resulting ordering depends on the set of languages under study and on whether automatic IPA transcription is being used or not. For each of the three language subsets under study, we can obtain character matrices either with or without using epitran. This results in $6$ possible combinations. For each of them, we construct a character matrix based on the $5000$ synsets with the most information available. \\
 
The second synset selection approach uses predefined concept lists. Here, we use the \textit{Swadesh-100-List} \cite{swadesh55} and \textit{core-wordnet}. Core-Wordnet is a list that has been extracted from WordNet and that contains the $5000$ most common nouns, verbs, and adjectives in the English language \cite{jordan06}. Working with concept lists might lead to the use of synsets with less data available. However, the lists contain more frequently used concepts, for which we expect to observe less horizontal transfer events \cite{haspelmath03} and thus an improved phylogenetic signal in the final character matrix. \\
\newpage
To obtain a synset for each concept in a given list, we query BabelNet with the corresponding English term. A query returns all synsets, for which the word has been specified as sense in that language. This poses the challenge of selecting one. In BabelNet, senses representing a part of the basic lexicon of a language are manually tagged as \textit{key sense}\cite{cecconi24-private}. If at least one synset is returned where the English main sense is marked as the key sense, we discard all synsets where this is not the case. From the remaining synsets, we then select that synset for which a main sense is present in most languages. 
Note that the datasets resulting from the synsets that have been selected based on the Swadesh-100-List are not converted into character matrices, as they contain too few entries per language to allow for a meaningful cognate clustering \cite{list14}. Instead, we use them to assess the sparsity of the data available in BabelNet in \cref{sec:compeleteness}. 

\subsubsection{Automatic IPA Transcription and Tokenization}
\label{sec:transcription}
The proportion of IPA transcriptions available in BabelNet is prohibitively small (see \cref{fig:filter}). In order to build more densely populated character matrices, we need to automatically transcribe orthographic words into IPA using epitran \cite{mortensen18}. We obtain the epitran instance for a certain language via its ISO code which is also used in BabelNet for language identification. For languages with more than one script, we use epitran's backoff class. \\
In addition, the IPA strings must be tokenized to apply cognate clustering. For this step, we use the ipatok (\url{https://github.com/pavelsof/ipatok}) python package. Both steps exhibit a high error rate. We experimentally determine this error rate via a reverse engineering approach in \cref{sec:reverse}).\\

\subsubsection{Cognate Clustering}
\label{sec:cognate-clustering}
Various tools for cognate clustering are described in the literature, which pursue different approaches \cite{akavarapu24,rama19,jaeger17}. We choose LexStat \cite{list12} because of its ease-of-use and the feasible result quality \cite{list12,akavarapu24}. Note that automatic cognate clustering also introduces errors. As we consider data sparsity to be the major challenge in character matrix generation, we omit comparing different cognate clustering methods here. Finally, we construct binary character matrices as described in \cite{haeuser24} for the resulting cognate datasets.\\

\subsection{Reverse Engineering for IPA Transcription and Tokenization}
\label{sec:reverse}
With the experiment presented in this section, we aim to quantify the induced error by automatic IPA transcription and tokenization via a reverse engineering approach.
Initially, we evaluate the error rate induced by automatic tokenization with ipatok. We extract IPA transcriptions as well as their tokenizations from lexibank-analysed \cite{list22} and from NorthEuraLex \cite{dellert19}. We tokenize each IPA transcription using ipatok and compare the result with the existing tokenization. We observe an error rate of $39.3\%$ for NorthEuraLex and of $54.2\%$ for lexibank-analysed. An alternative implementation for IPA tokenization is also available in the lingpy \cite{list24} tool. However, the resulting error rates are slightly higher ($50.0\%$ for NorthEuraLex, $62.4\%$ for lexibank-analysed). Therefore, we use ipatok in the following.\\
\newpage
For assessing the quality of epitran-based transcriptions, we require pairs of orthographic words and their corresponding IPA transcriptions. Thus, we only work with data from NorthEura- Lex, as lexibank-analysed does not contain orthographic words. We consider the results for the different languages separately. The error rates $e_1$ are provided in \cref{tab:epitran-error-rates}. We observe varying, yet overall excessively high error rates. \\
For cognate clustering, the tokens are converted into dolgo sound classes \cite{list12, dolgopolsky64}. Thus, a substitution error in the IPA sequence does not affect the final character matrix as long as an incorrect token belongs to the same sound class as the correct one. Therefore, we reassess the epitran transcriptions with respect to this observation. We tokenize both, the epitran transcriptions, and the supplied transcriptions using ipatok. Thereby, we aim to abstract from errors resulting from  automatic tokenization. We compare the dolgo sound classes of the tokens from the epitran transcription to those obtained for the existing transcription. We again consider the languages separately (see $e_2$ in \cref{tab:epitran-error-rates}). For many languages, the error rates in this study are lower, suggesting that the errors induced by automatic IPA transcription do not affect the final result. On the other hand, languages with high error rates are still present. \\

\begin{table}
\centering
\begin{tabular}{l|r|r}
\hline
 glottocode   & $e_1$   & $e_2$   \\
\hline
 hind1269     & $74.2 \%$    & $34.6 \%$          \\
 stan1288     & $51.57 \%$   & $17.52 \%$         \\
 nucl1301     & $56.12 \%$   & $16.33 \%$         \\
 stan1293     & $85.5 \%$    & $52.89 \%$         \\
 stan1289     & $92.06 \%$   & $47.22 \%$         \\
 croa1245     & $99.77 \%$   & $0.94 \%$          \\
 kaza1248     & $82.47 \%$   & $60.9 \%$          \\
 czec1258     & $60.97 \%$   & $7.7 \%$           \\
 ukra1253     & $30.31 \%$   & $8.04 \%$          \\
 avar1256     & $70.55 \%$   & $64.68 \%$         \\
 telu1262     & $96.08 \%$   & $42.41 \%$         \\
 russ1263     & $100.0 \%$   & $9.64 \%$          \\
 mala1464     & $90.92 \%$   & $56.91 \%$         \\
 poli1260     & $50.8 \%$    & $24.64 \%$         \\
 stan1295     & $74.51 \%$   & $16.44 \%$         \\
 beng1280     & $92.12 \%$   & $42.9 \%$          \\
 swed1254     & $82.56 \%$   & $13.5 \%$          \\
 dutc1256     & $69.57 \%$   & $53.0 \%$          \\
 alba1267     & $25.85 \%$   & $1.19 \%$          \\
 hung1274     & $1.38 \%$    & $0.65 \%$          \\
 port1283     & $97.91 \%$   & $35.02 \%$         \\
 nort2641     & $95.32 \%$   & $10.25 \%$         \\
 roma1327     & $34.48 \%$   & $26.99 \%$         \\
 tami1289     & $93.03 \%$   & $54.17 \%$         \\
 stan1290     & $88.07 \%$   & $47.96 \%$         \\
 ital1282     & $28.18 \%$   & $17.6 \%$          \\
 mand1415     & $100.0 \%$   & $38.81 \%$         \\
\hline
\end{tabular}
\label{tab:epitran-error-rates}
\caption{Error rates of epitran-based IPA transcriptions for NorthEuraLex. To determine the error rate $e_1$, a transcription obtained from epitran is considered as being correct if and only if it is identical to the transcription provided in NorthEuraLex. To obtain the error rate $e_2$, we consider an epitran-based transcription to be correct if it corresponds to the same dolgo sound classes as the transcription from NorthEuraLex, even if the two transcriptions are not strictly identical.}
\end{table}

We conduct an additional study to examine the effect of the automatic IPA transcription and tokenization on the phylogenetic signal of the resulting dataset. To this end, we compare three versions of the NorthEuraLex dataset. In the original version, we use the available IPA transcriptions and tokenizations. In the second version, we also use the available IPA transcription but tokenize it automatically with ipatok. Using both ipatok and epitran, we obtain a third version in which the IPA transcriptions {\em and} their tokenizations are created in an automated manner. For each version, we conduct cognate clustering and determine the corresponding binary character matrix as described in \cref{sec:cognate-clustering}. On each of these character matrices, we execute $20$ tree searches using the default tree search of RAxML-NG v. 1.2.0 ($10$ searches starting from random trees and $10$ searches starting from randomized stepwise addition order parsimony trees). We consider the best-scoring trees resulting from these inferences and determine their GQ distances to the gold standard Glottolog tree. Further, we determine the Pythia ground truth difficulty scores for character matrices corresponding to the different versions of the NorthEuraLex dataset. These scores quantify the difficulty of a phylogenetic inference on a dataset ranging from $0$ (easy) to $1$ (hopeless) \cite{haag25}. The results are provided in \cref{tab:signal-northeuralex}. We observe that both, the GQ distance to the gold standard, and the ground truth difficulty, are higher if ipatok and/or epitran are used. This indicates that automatic IPA transcription and tokenization yield datasets with a weaker phylogenetic signal.\\

\begin{table}[]
    \centering
    \begin{tabular}{l|r|r}
    \hline
        data &  GQ distance & ground truth difficulty\\
         \hline
        original & $0.317$ & $0.757$\\
        \hline
        ipatok & $0.366$ & $0.878$\\
        \hline
        epitran + ipatok & $0.401$ & $0.861$\\
        \hline
    \end{tabular}
    \caption{The table shows the impact of the automatic IPA transcription and tokenization on the phylogenetic signal of the NorthEuraLex dataset. We compare three versions of the NorthEuraLex dataset. (Original version using provided IPA transcriptions and tokenizations, ipatok version using the provided IPA transcription and automated tokenization, ipatok + epitran version using automated IPA transcription and tokenization). For each version, the table comprises the GQ distances between the best-scoring Maximum Likelihood tree and the gold standard tree from Glottolog as well as the Pythia difficulty score. Both are higher if ipatok and/or epitran are used, indicating that it leads to a weaker phylogenetic signal.}    
    \label{tab:signal-northeuralex}
\end{table}

Note that our reverse engineering assessment only takes the quality of epitran for languages for which data {\em are} available in NorthEuraLex into account. Numerous other languages are supported, but the quality of the corresponding transcriptions is not examined here.

\newpage

\subsection{Evaluation}
\label{sec:evaluation}
In this section, we evaluate the datasets we extracted from BabelNet. In \cref{sec:compeleteness} we compare character matrices extracted from BabelNet to manually constructed ones with respect to their density, that is, the completeness of the data. In \cref{sec:signal} we analyze the results of ML tree inferences on the character matrices obtained from BabelNet and we compute their Pythia difficulty scores in order to assess the phylogenetic signal contained in the data.
\subsubsection{Completeness of the Data}
\label{sec:compeleteness}
As described in \cref{sec:synset-selection}, we explore two different approaches for selecting the synsets we include in our character matrices. In the first approach, we use the $5000$ synsets for which IPA transcriptions are available in most languages. The properties of the resulting datasets are given in \cref{tab:data-filter}. 
The table also provides the \textit{average mutual coverage} (AMC). For a multilingual wordlist, the AMC is defined as the average number of concepts that are shared by all language pairs divided by the overall number of concepts. Hence, AMC measures the concept overlap \cite{list18-clics}. In the following we use the AMC implementation from the LingPy software package \cite{list24}.
Considering the number of synsets, the question arises as to why less than $5000$ synsets are contained in the datasets. Due to the data quality, not all main senses that were counted in the statistics described in \cref{sec:synset-selection} can ultimately be taken into account. This is mainly because the IPA transcriptions contain symbols that are not part of the official alphabet and can therefore not be processed in the following steps. This applies both, to IPA transcriptions from BabelNet, and to those obtained via epitran.\\
As expected, we obtain the densest dataset for the \textit{dense} languages in conjunction with the use of epitran. This is also reflected by the comparatively high AMC of $0.490$. In this case, a sense is available on average for more than every second synset of the languages under consideration. In the remaining constellations, the data become sparser. Even in the densest character matrix, however, the coverage is still substantially worse than in manually assembled character matrices for phylogenetic inferences which have an AMC $> 0.85$ \cite{haeuser25}.

\begin{table*}[]
    \centering
    \begin{tabular*}{\textwidth}{r l|r|r|r|r|r}
    \hline
         &&  \#langs. & \#synsets & \#langs. per synset & \#synsets per lang. & AMC\\
         \hline
        all& & $101$ & $3653$ & $2.3$ & $83.3$ & $0.001$\\
        all&+ epitran & $136$ & $4790$ & $50.3$ & $1771.3$ & $0.157$\\
        \hline
        dense& & $44$ & $3018$& $1.9$ & $127.4$ & $0.001$\\
        dense&+ epitran & $77$ & $4778$ & $50.2$ & $3113.6$ & $0.490$\\
        \hline
        iecor& & $45$ & $3250$ & $2.0$ & $145.0$ & $0.002$\\
        iecor&+ epitran & $52$ & $4727$ & $16.0$ & $1450.8$ & $0.103$\\
        \hline
    \end{tabular*}
    \caption{Properties of datasets obtained from the $5000$ synsets with most data available.}
    \label{tab:data-filter}
\end{table*}

\newpage
For the second approach, we select synsets that are based on the core-wordnet concept list. The properties of the resulting datasets are given in \cref{tab:data-core-wordnet}. In general, the results are similar to what we observe for the first synset selection approach. However, using epitran yields a less pronounced improvement regarding the number of synsets with senses available per language. Also, the AMC is substantially lower.

\begin{table*}[]
    \centering
    \begin{tabular*}{\textwidth}{r l|r|r|r|r|r}
    \hline
         &&  \#langs. & \#synsets & \#langs. per synset & \#synsets per lang. & AMC\\
         \hline
        all& & $95$ & $2867$ & $2.3$ & $69.3$ & $0.001$\\
        all&+ epitran & $129$ & $4848$ & $18.8$ & $705.7$ & $0.035$\\
        \hline
        dense& & $42$ & $2369$& $1.9$ & $101.4$ & $0.002$\\
        dense&+ epitran & $77$ & $4835$ & $18.3$ & $1150.7$ & $0.096$\\
        \hline
        iecor& & $43$ & $2523$ & $1.9$ & $114.0$ & $0.002$\\
        iecor&+ epitran & $50$ & $4816$ & $8.8$ & $842.8$ & $0.036$\\
        \hline
    \end{tabular*}
    \caption{Properties of datasets obtained based on the core-wordnet conceptlist.}
    \label{tab:data-core-wordnet}
\end{table*}

\begin{figure}[h!]
    \centering
      \begin{subfigure}{0.8\textwidth}
        \includegraphics[width=\textwidth]{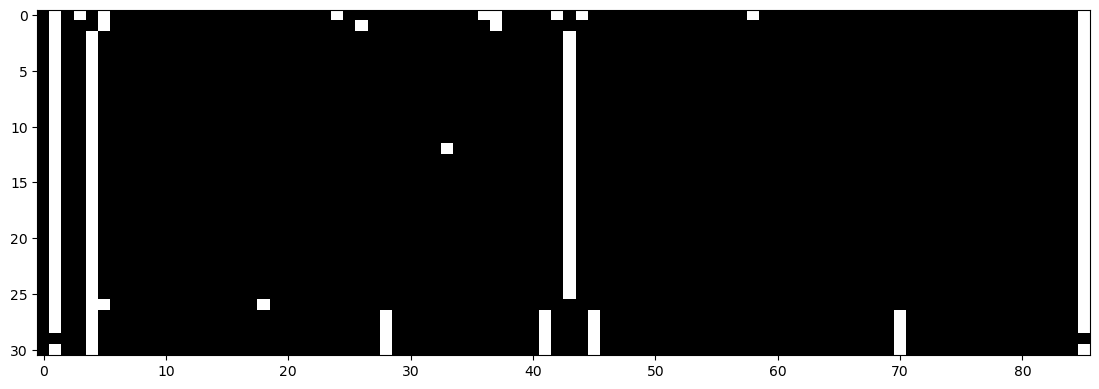}
          \caption{\textit{iecor} languages}
          \label{fig:iecor-lexibank-sparsity}
      \end{subfigure}
      \begin{subfigure}{0.8\textwidth}
        \includegraphics[width=\textwidth]{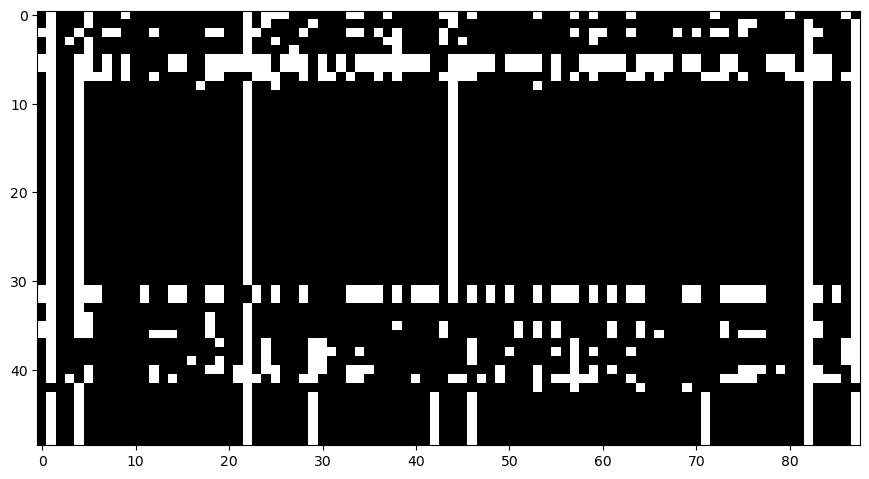}
          \caption{\textit{dense} languages}
          \label{fig:main-lexibank-sparsity}
      \end{subfigure}
\label{fig:lexibank-sparsity}
\caption{Data availability for the Swadesh-100 list in lexibank-analysed}
\end{figure}

\begin{figure}[h!]
    \centering
      \begin{subfigure}{0.8\textwidth}
        \includegraphics[width=\textwidth]{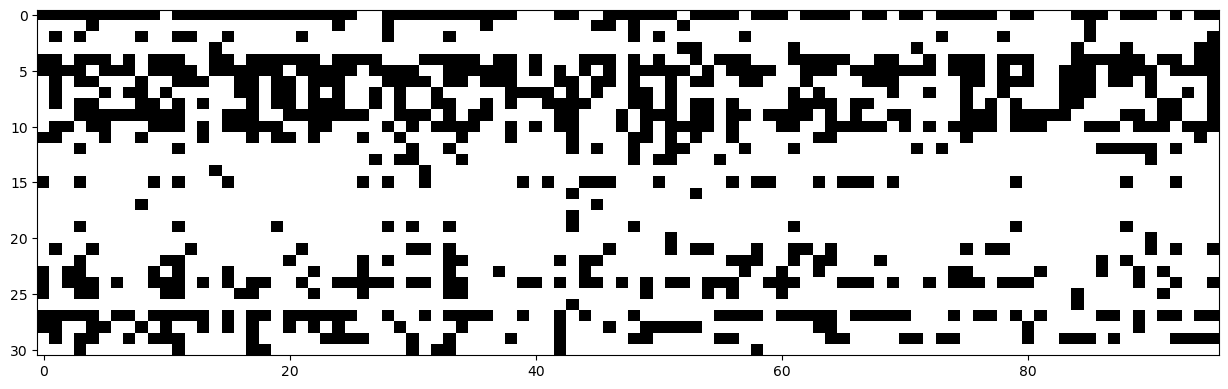}
          \caption{\textit{iecor} languages}
          \label{fig:iecor-babelnet-sparsity}
      \end{subfigure}
      \begin{subfigure}{0.8\textwidth}
        \includegraphics[width=\textwidth]{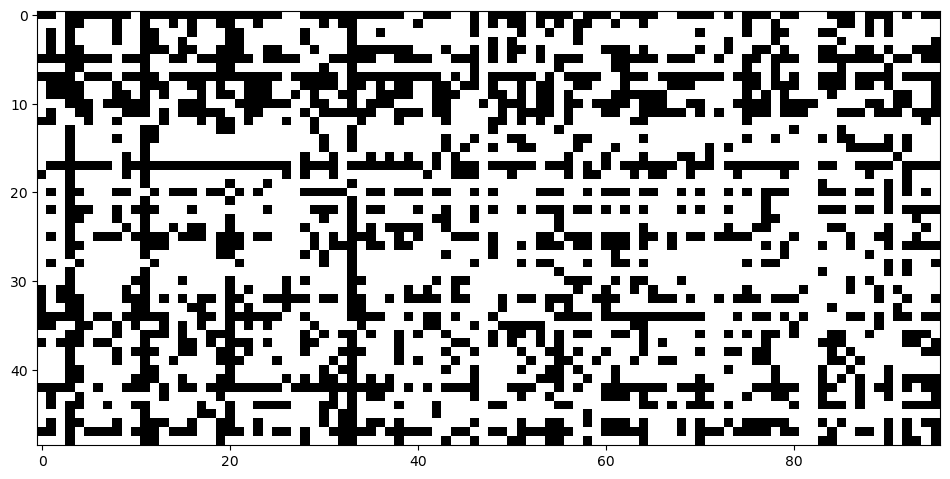}
          \caption{\textit{dense} languages}
          \label{fig:main-babelnet-sparsity}
      \end{subfigure}
\label{fig:babelnet-sparsity}
\caption{Data availability for the Swadesh-100 list in BabelNet, with epitran used for automatic IPA transcription}
\end{figure}

For a better assessment of the amount of available data, we conduct a comparison to the density of manually assembled datasets. The manual data collection process often relies on the Swadesh-100-List. Therefore, we extract datasets from BabelNet that are also based on this concept list. For an intuitive visualization of the available data, we use so-called sparsity plots. These plots have the structure of a two-dimensional matrix where the rows correspond to the languages and the columns correspond to the concepts contained in the dataset under study. If there is a word specified for a certain language-concept pair, the respective matrix cell is colored black, otherwise it is left blank. \cref{fig:lexibank-sparsity} illustrates the amount of manually collected data available in the meta-dataset lexibank-analysed \cite{list22} for the \textit{iecor} languages (\cref{fig:iecor-lexibank-sparsity}) and for the \textit{dense} languages (\cref{fig:main-lexibank-sparsity}). In contrast to that, \cref{fig:babelnet-sparsity} provides sparsity plots showing the amount of data available in BabelNet in combination with automatic IPA transcription via epitran. Note that the plots are restricted to the languages, for which data are provided in {\em both} lexibank-analysed as well as in the BabelNet extract.

We observe a substantial difference in dataset density. While for manually collected data, the wordlists are almost fully occupied, there is a large proportion of missing entries in the corresponding datasets retrieved from BabelNet. The only approach to attain comparable results with BabelNet data is to include more concepts and building larger datasets. To this end, we use up to $5000$ instead of only $100$ concepts. In the following section, we evaluate the performance of phylogenetic inferences on these resulting large BabelNet character matrices.

\subsubsection{Phylogenetic Signal}
\label{sec:signal}
We use the two different synset selection approaches (see \cref{sec:synset-selection} and apply them to three different language sets (\textit{all}, \textit{dense} and \textit{iecor}). We further construct datasets with and without automatic epitran-based IPA transcription. This yields $12$ different character matrices. On each of them, we execute $20$ independent Maximum Likelihood (ML) tree searches. We again use the default RAxML-NG tree search setting ($10$ searches starting from random trees and $10$ searches starting from randomized stepwise addition order parsimony trees). We apply the BIN+G model of binary character substitution to accommodate among site rate heterogeneity via the $\Gamma$-model. \\
We assess a tree inferred on a character matrix by comparing it to the corresponding gold standard tree for the respective languages. We extract this gold standard tree from the manually constructed tree published in the \textit{Glottolog} database \cite{hammarstroem22}. To compare an inferred ML tree to the gold standard, we use the \textit{generalized quartet (GQ) distance} \cite{pompei11}. This metric has the advantage that it yields a distance of $0$ if there are no contradictions between the inferred tree and the gold standard tree. This even holds if the gold standard tree contains polytomies, which is not uncommon for Glottolog trees. To calculate the GQ distance, one extracts all possible quartets of tips induced by the tree. For each quartet, one then determines the topology of the induced $4$-tip subtree. When comparing two trees, the GQ distance reflects the proportion of quartets for which the induced subtrees exhibit distinct topologies. \\
\cref{tab:results-filter} and \cref{tab:results-core-wordnet} show the GQ distances of the best-scoring trees to the Glottolog gold standard tree. Running inferences on the character matrices constructed without automatic IPA transcription yields trees with GQ distances $> 0.5$ to the reference. The usage of epitran yields trees that are closer to the gold standard and therefore improves results. However, none of the inferred trees attains a GQ distance substantially below $0.4$. This indicates that the differences are still substantial, in particular when considering the fact that inferring a tree on the character matrix representing the manually assembled iecor database, we obtain a GQ distance of $0.024$ to the gold standard.\\
The tables also show the Pythia ground truth difficulty scores \cite{haag22} for the character matrices, which range from $0.620$ to $0.926$ indicating a weak phylogenetic signal. 
Overall, these observations show that the constructed character matrices are not suitable for phylogenetic inference.

\begin{table}[]
    \centering
    \begin{tabular}{r l|r|r}
    \hline
         &&  GQ distance & ground truth difficulty\\
         \hline
        all & & $0.628$ & $0.926$\\
        all & + epitran & $0.554$ & $0.764$\\
        \hline
        dense & & $0.593$ & $0.824$\\
        dense & + epitran & $0.482$ & $0.620$\\
        \hline
        iecor & & $0.538$ & $0.823$\\
        iecor & + epitran & $0.400$ & $0.704$\\
        \hline
    \end{tabular}
    \caption{Results obtained for the character matrices constructed based on the $5000$ synsets with most data available}    
    \label{tab:results-filter}
\end{table}

\begin{table}[]
    \centering
    \begin{tabular}{r l|r|r}
    \hline
         &&  GQ distance & ground truth difficulty\\
         \hline
        all & & $0.604$ & $0.916$\\
        all & + epitran & $0.451$ & $0.818$\\
        \hline
        dense & & $0.641$ & $0.816$\\
        dense & + epitran & $0.446$ & $0.693$\\
        \hline
        iecor & & $0.647$ & $0.834$\\
        iecor & + epitran & $ 0.394$ & $0.828$\\
        \hline
    \end{tabular}
    \caption{Results obtained for the character matrices constructed based on the core-wordnet conceptlist}    
    \label{tab:results-core-wordnet}
\end{table}

\section{Conclusion and Discussion}
\label{sec:conclusion}
Initially, we motivated our work by the need for larger cognate datasets to benefit from recent advances in phylogenetics by applying sophisticated models and machine learning-based techniques.
In \cref{sec:sources} we assessed numerous multilingual resources and explained why most of them are not suitable for automatically extracting data for downstream phylogenetic inference. We selected the multilingual encyclopedic dictionary BabelNet to automatically generate character matrices (see \cref{sec:babelnet}). While BabelNet appears to be a promising resource at first sight since it contains data for over $600$ languages, we were only able to obtain sufficiently dense matrices for up to $132$ languages. Based on the results from the ML tree inferences and from the character matrices' Pythia difficulty scores, we concluded that the automatically extracted character matrices from BabelNet are not suitable for phylogenetic inference. \\
We were not able to compensate for the disadvantage of automated data collection, that is, poorer quality, by means of a comparatively seamless acquisition of more data. One reason for this is the general data sparsity, especially for low-resource languages (see \cref{sec:sources}). Our work shows that this still constitutes an unresolved challenge, despite the fact that multilingual resources are growing in number and size. Some errors directly result from the fact that we automatically query BabelNet. To avoid these, it would be necessary to assess whether the retrieved words adequately describe the requested concepts. Another reason for the low quality of the final character matrices is the introduction of errors by automatic IPA transcription and tokenization (see \cref{sec:reverse}) which we quantify via reverse-engineering. IPA transcription and tokenization could be improved, if both steps were carried out simultaneously. Automatic cognate clustering constitutes another likely source of error. Investigating different parameter configurations for the clustering algorithm could improve the results of this step. To generate larger cognate datasets, we extended the underlying concept lists by using less fundamental concepts. However, these concepts are more susceptible to horizontal transfer \cite{haspelmath03}, which may also contribute to the poor signal we observed for the extracted character matrices. 
Of all candidate data sources listed in \cref{sec:sources}, we chose BabelNet for character matrix extraction as it appeared to be the most suitable for this purpose. When compiling datasets from a resource other than BabelNet, we might encounter analogous, potentially more pronounced, challenges as with BabelNet. We therefore expect the resulting character matrices to be of even poorer quality. \\
To the best of our knowledge, there currently exists no feasible approach to generate larger cognate datasets. This means that numerous recent advances in computational molecular phylogenetics, that is, more sophisticated models and machine learning-based approaches, can currently not be applied to cognate data, and we also advise against doing so. To move forward, one needs to investigate fundamentally distinct approaches to acquire language data for phylogenetic inference, such as, for instance, applying machine learning methods for character matrix extraction from sound recordings \cite{jaeger24}.

\section*{Acknowledgement}
Luise Häuser and Alexandros Stamatakis are financially supported by the Klaus Tschira Foundation, and by the European Union (EU) under Grant Agreement No 101087081 (CompBiodiv-GR).
\\
We would like to thank Gerhard Jäger and Mattis List for their support with their detailed expertise in the field of historical linguistics and to Michael Strube for his numerous suggestions on potential data sources.\\

\includegraphics[width=0.49\textwidth]{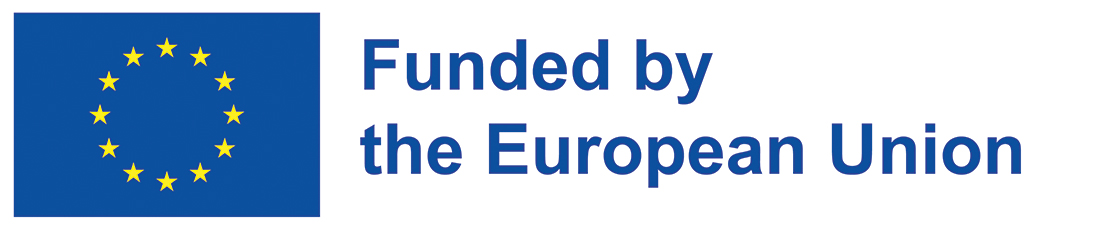}\vspace{20cm}

\newpage
\bibliography{main}

\begin{thebibliography}{66}
\providecommand{\natexlab}[1]{#1}
\providecommand{\url}[1]{\texttt{#1}}
\expandafter\ifx\csname urlstyle\endcsname\relax
  \providecommand{\doi}[1]{doi: #1}\else
  \providecommand{\doi}{doi: \begingroup \urlstyle{rm}\Url}\fi

\bibitem[Adebara et~al.(2023)Adebara, Elmadany, Abdul-Mageed, and Inciarte]{adebara23}
Ife Adebara, AbdelRahim Elmadany, Muhammad Abdul-Mageed, and Alcides~Alcoba Inciarte.
\newblock Serengeti: Massively multilingual language models for africa, 2023.

\bibitem[Adelani et~al.(2024)Adelani, Liu, Shen, Vassilyev, Alabi, Mao, Gao, and Lee]{adelani24}
David Adelani, Hannah Liu, Xiaoyu Shen, Nikita Vassilyev, Jesujoba Alabi, Yanke Mao, Haonan Gao, and En-Shiun Lee.
\newblock {SIB}-200: A simple, inclusive, and big evaluation dataset for topic classification in 200+ languages and dialects.
\newblock In Yvette Graham and Matthew Purver, editors, \emph{Proceedings of the 18th Conference of the European Chapter of the Association for Computational Linguistics (Volume 1: Long Papers)}, pages 226--245, St. Julian{'}s, Malta, March 2024. Association for Computational Linguistics.
\newblock URL \url{https://aclanthology.org/2024.eacl-long.14}.

\bibitem[Akavarapu and Bhattacharya(2024{\natexlab{a}})]{akavarapu24}
V.~S. D. S.~Mahesh Akavarapu and Arnab Bhattacharya.
\newblock Automated cognate detection as a supervised link prediction task with cognate transformer, 2024{\natexlab{a}}.

\bibitem[Akavarapu and Bhattacharya(2024{\natexlab{b}})]{akavarapu24-likelihood}
V.S.D.S.Mahesh Akavarapu and Arnab Bhattacharya.
\newblock A likelihood ratio test of genetic relationship among languages.
\newblock In Kevin Duh, Helena Gomez, and Steven Bethard, editors, \emph{Proceedings of the 2024 Conference of the North American Chapter of the Association for Computational Linguistics: Human Language Technologies (Volume 1: Long Papers)}, pages 2559--2570, Mexico City, Mexico, June 2024{\natexlab{b}}. Association for Computational Linguistics.
\newblock \doi{10.18653/v1/2024.naacl-long.141}.
\newblock URL \url{https://aclanthology.org/2024.naacl-long.141}.

\bibitem[Almeida and Xexéo(2023)]{almeida23}
Felipe Almeida and Geraldo Xexéo.
\newblock Word embeddings: A survey, 2023.
\newblock URL \url{https://arxiv.org/abs/1901.09069}.

\bibitem[Azouri et~al.(2021)Azouri, Abadi, Mansour, Mayrose, and Pupko]{azouri21}
Dana Azouri, Shiran Abadi, Yishay Mansour, Itay Mayrose, and Tal Pupko.
\newblock {Harnessing machine learning to guide phylogenetic-tree search algorithms}.
\newblock \emph{Nature Communications}, 12\penalty0 (1):\penalty0 1--9, December 2021.
\newblock \doi{10.1038/s41467-021-22073-}.
\newblock URL \url{https://ideas.repec.org/a/nat/natcom/v12y2021i1d10.1038_s41467-021-22073-8.html}.

\bibitem[Bandarkar et~al.(2023)Bandarkar, Liang, Muller, Artetxe, Shukla, Husa, Goyal, Krishnan, Zettlemoyer, and Khabsa]{bandarkar23}
Lucas Bandarkar, Davis Liang, Benjamin Muller, Mikel Artetxe, Satya~Narayan Shukla, Donald Husa, Naman Goyal, Abhinandan Krishnan, Luke Zettlemoyer, and Madian Khabsa.
\newblock The belebele benchmark: a parallel reading comprehension dataset in 122 language variants, 2023.

\bibitem[Bond et~al.(2016)Bond, Vossen, McCrae, and Fellbaum]{bond16}
Francis Bond, Piek Vossen, John McCrae, and Christiane Fellbaum.
\newblock {CILI}: the collaborative interlingual index.
\newblock In Christiane Fellbaum, Piek Vossen, Verginica~Barbu Mititelu, and Corina Forascu, editors, \emph{Proceedings of the 8th Global WordNet Conference (GWC)}, pages 50--57, Bucharest, Romania, 27--30 January 2016. Global Wordnet Association.
\newblock URL \url{https://aclanthology.org/2016.gwc-1.9}.

\bibitem[Boyd-Graber et~al.(2006)Boyd-Graber, Fellbaum, Osherson, and Schapire]{jordan06}
Jordan Boyd-Graber, Christiane Fellbaum, Daniel Osherson, and Robert Schapire.
\newblock Adding dense, weighted connections to {WordNet}.
\newblock \emph{GWC 2006: 3rd International Global WordNet Conference, Proceedings}, 01 2006.

\bibitem[Cecconi()]{cecconi24-private}
Francesco Cecconi.
\newblock Personal Communication.

\bibitem[Cieri et~al.(2016)Cieri, Maxwell, Strassel, and Tracey]{cieri16}
Christopher Cieri, Mike Maxwell, Stephanie Strassel, and Jennifer Tracey.
\newblock Selection criteria for low resource language programs.
\newblock In Nicoletta Calzolari, Khalid Choukri, Thierry Declerck, Sara Goggi, Marko Grobelnik, Bente Maegaard, Joseph Mariani, Helene Mazo, Asuncion Moreno, Jan Odijk, and Stelios Piperidis, editors, \emph{Proceedings of the Tenth International Conference on Language Resources and Evaluation ({LREC}'16)}, pages 4543--4549, Portoro{\v{z}}, Slovenia, May 2016. European Language Resources Association (ELRA).
\newblock URL \url{https://aclanthology.org/L16-1720}.

\bibitem[Dellert et~al.(2019)Dellert, Daneyko, and Münch]{dellert19}
J.~Dellert, T.~Daneyko, and A.~et~al. Münch.
\newblock Lang resources \& evaluation, 2019.

\bibitem[Devlin et~al.(2019)Devlin, Chang, Lee, and Toutanova]{devlin19}
Jacob Devlin, Ming-Wei Chang, Kenton Lee, and Kristina Toutanova.
\newblock Bert: Pre-training of deep bidirectional transformers for language understanding, 2019.
\newblock URL \url{https://arxiv.org/abs/1810.04805}.

\bibitem[Dolgopolsky(1964)]{dolgopolsky64}
Aron~B. Dolgopolsky.
\newblock Gipoteza drevnejšego rodstva jazykovych semej severnoj evrazii s verojatnostej točky zrenija, 1964.

\bibitem[Dunn(2013)]{dunn13}
Michael Dunn.
\newblock Language phylogenies, 08 2013.

\bibitem[Evans et~al.(2006)Evans, Ringe, and Warnow]{evans06}
Steven Evans, Don Ringe, and Tandy Warnow.
\newblock Inference of divergence times as a statistical inverse problem.
\newblock \emph{Phylogenetic methods and the prehistory of languages}, 01 2006.

\bibitem[FitzGerald et~al.(2023)FitzGerald, Hench, Peris, Mackie, Rottmann, Sanchez, Nash, Urbach, Kakarala, Singh, Ranganath, Crist, Britan, Leeuwis, Tur, and Natarajan]{fitzgerald23}
Jack FitzGerald, Christopher Hench, Charith Peris, Scott Mackie, Kay Rottmann, Ana Sanchez, Aaron Nash, Liam Urbach, Vishesh Kakarala, Richa Singh, Swetha Ranganath, Laurie Crist, Misha Britan, Wouter Leeuwis, Gokhan Tur, and Prem Natarajan.
\newblock {MASSIVE}: A 1{M}-example multilingual natural language understanding dataset with 51 typologically-diverse languages.
\newblock In Anna Rogers, Jordan Boyd-Graber, and Naoaki Okazaki, editors, \emph{Proceedings of the 61st Annual Meeting of the Association for Computational Linguistics (Volume 1: Long Papers)}, pages 4277--4302, Toronto, Canada, July 2023. Association for Computational Linguistics.
\newblock \doi{10.18653/v1/2023.acl-long.235}.
\newblock URL \url{https://aclanthology.org/2023.acl-long.235}.

\bibitem[Guha et~al.(2015)Guha, Brickley, and MacBeth]{guha15}
Ramanathan Guha, Dan Brickley, and Steve MacBeth.
\newblock Schema.org: Evolution of structured data on the web.
\newblock \emph{Queue}, 13:\penalty0 10--37, 11 2015.
\newblock \doi{10.1145/2857274.2857276}.

\bibitem[Haag and Stamatakis(2025)]{haag25}
Julia Haag and Alexandros Stamatakis.
\newblock Pythia 2.0: New data, new prediction model, new features.
\newblock \emph{bioRxiv}, 2025.
\newblock \doi{10.1101/2025.03.25.645182}.
\newblock URL \url{https://www.biorxiv.org/content/early/2025/03/28/2025.03.25.645182}.

\bibitem[Haag et~al.(2022)Haag, H{\"o}hler, Bettisworth, and Stamatakis]{haag22}
Julia Haag, Dimitri H{\"o}hler, Ben Bettisworth, and Alexandros Stamatakis.
\newblock From easy to hopeless - predicting the difficulty of phylogenetic analyses.
\newblock \emph{bioRxiv}, 2022.
\newblock \doi{10.1101/2022.06.20.496790}.

\bibitem[Hammarström et~al.(2022)Hammarström, Forkel, Haspelmath, and Bank]{hammarstroem22}
Harald Hammarström, Robert Forkel, Martin Haspelmath, and Sebastian Bank.
\newblock Glottolog 4.7, 2022.
\newblock URL \url{http://glottolog.org}.

\bibitem[Haspelmath(2003)]{haspelmath03}
Martin Haspelmath, 2003.

\bibitem[Heggarty(2023)]{heggarty23}
Paul et~al. Heggarty.
\newblock Language trees with sampled ancestors support a hybrid model for the origin of indo-european languages.
\newblock \emph{Science}, 381\penalty0 (6656), 2023.
\newblock \doi{10.1126/science.abg0818}.

\bibitem[Heinzerling and Strube(2018)]{heinzerling18}
Benjamin Heinzerling and Michael Strube.
\newblock {BPEmb: Tokenization-free Pre-trained Subword Embeddings in 275 Languages}.
\newblock In Nicoletta Calzolari~(Conference chair), Khalid Choukri, Christopher Cieri, Thierry Declerck, Sara Goggi, Koiti Hasida, Hitoshi Isahara, Bente Maegaard, Joseph Mariani, Hélène Mazo, Asuncion Moreno, Jan Odijk, Stelios Piperidis, and Takenobu Tokunaga, editors, \emph{Proceedings of the Eleventh International Conference on Language Resources and Evaluation (LREC 2018)}, Miyazaki, Japan, May 7-12, 2018 2018. European Language Resources Association (ELRA).
\newblock ISBN 979-10-95546-00-9.

\bibitem[Hu et~al.(2020)Hu, Ruder, Siddhant, Neubig, Firat, and Johnson]{hu20}
Junjie Hu, Sebastian Ruder, Aditya Siddhant, Graham Neubig, Orhan Firat, and Melvin Johnson.
\newblock Xtreme: A massively multilingual multi-task benchmark for evaluating cross-lingual generalization, 2020.

\bibitem[Häuser and List()]{haeuser25}
Luise Häuser and Johann-Mattis List.
\newblock Lexibench: Towards an improved collection of benchmark data for computational historical linguistics.
\newblock URL \url{https://calc.hypotheses.org/8227}.

\bibitem[Häuser et~al.()Häuser, Jäger, and Stamatakis]{haeuser25-3}
Luise Häuser, Gerhard Jäger, and Alexandros Stamatakis.
\newblock A systematic exploration of current limitations of cognate-based phylogenetic inference.

\bibitem[Häuser et~al.(2024{\natexlab{a}})Häuser, Jäger, Rama, List, and Stamatakis]{haeuser2024-cp}
Luise Häuser, Gerhard Jäger, Taraka Rama, Johann-Mattis List, and Alexandros Stamatakis.
\newblock Are sounds sound for phylogenetic reconstruction?, 2024{\natexlab{a}}.
\newblock URL \url{https://arxiv.org/abs/2402.02807}.

\bibitem[Häuser et~al.(2024{\natexlab{b}})Häuser, Jäger, and Stamatakis]{haeuser24}
Luise Häuser, Gerhard Jäger, and Alexandros Stamatakis.
\newblock Computational approaches for integrating out subjectivity in cognate synonym selection, 2024{\natexlab{b}}.
\newblock URL \url{https://arxiv.org/abs/2404.19328}.

\bibitem[ImaniGooghari et~al.(2023)ImaniGooghari, Lin, Kargaran, Severini, Jalili~Sabet, Kassner, Ma, Schmid, Martins, Yvon, and Schütze]{googhari23}
Ayyoob ImaniGooghari, Peiqin Lin, Amir~Hossein Kargaran, Silvia Severini, Masoud Jalili~Sabet, Nora Kassner, Chunlan Ma, Helmut Schmid, André Martins, François Yvon, and Hinrich Schütze.
\newblock Glot500: Scaling multilingual corpora and language models to 500 languages.
\newblock In \emph{Proceedings of the 61st Annual Meeting of the Association for Computational Linguistics (Volume 1: Long Papers)}. Association for Computational Linguistics, 2023.
\newblock \doi{10.18653/v1/2023.acl-long.61}.
\newblock URL \url{http://dx.doi.org/10.18653/v1/2023.acl-long.61}.

\bibitem[J{\"a}ger et~al.(2017)J{\"a}ger, List, and Sofroniev]{jaeger17}
Gerhard J{\"a}ger, Johann-Mattis List, and Pavel Sofroniev.
\newblock Using support vector machines and state-of-the-art algorithms for phonetic alignment to identify cognates in multi-lingual wordlists.
\newblock In Mirella Lapata, Phil Blunsom, and Alexander Koller, editors, \emph{Proceedings of the 15th Conference of the {E}uropean Chapter of the Association for Computational Linguistics: Volume 1, Long Papers}, pages 1205--1216, Valencia, Spain, April 2017. Association for Computational Linguistics.
\newblock URL \url{https://aclanthology.org/E17-1113}.

\bibitem[Jäger(2018)]{jaeger18}
Gerhard Jäger.
\newblock Global-scale phylogenetic linguistic inference from lexical resources.
\newblock \emph{Scientific Data}, 5, 10 2018.
\newblock \doi{10.1038/sdata.2018.189}.

\bibitem[Jäger(2024)]{jaeger24}
Gerhard Jäger.
\newblock Phylogenetic linguistic inference from acoustic speech data: Ideas for a novel research paradigm, 2024.
\newblock URL \url{https://profgerhard.de/slides/slides_zuerich2024.pdf}.

\bibitem[Kolipakam et~al.(2018)Kolipakam, Jordan, Dunn, Greenhill, Bouckaert, Gray, and Verkerk]{kolipakam18}
Vishnupriya Kolipakam, Fiona~M. Jordan, Michael Dunn, Simon~J. Greenhill, Remco Bouckaert, Russell~D. Gray, and Annemarie Verkerk.
\newblock A {Bayesian} phylogenetic study of the {Dravidian} language family.
\newblock \emph{Royal Society Open Science}, 5\penalty0 (171504):\penalty0 1--17, 2018.

\bibitem[Kontokostas et~al.(2012)Kontokostas, Bratsas, Auer, Hellmann, Antoniou, and Metakides]{kontokostas12}
Dimitris Kontokostas, Charalampos Bratsas, Sören Auer, Sebastian Hellmann, Ioannis Antoniou, and George Metakides.
\newblock Internationalization of linked data: The case of the greek dbpedia edition.
\newblock \emph{Journal of Web Semantics}, 15:\penalty0 51--61, 2012.
\newblock ISSN 1570-8268.
\newblock \doi{https://doi.org/10.1016/j.websem.2012.01.001}.
\newblock URL \url{https://www.sciencedirect.com/science/article/pii/S1570826812000030}.

\bibitem[Lehmann et~al.(2014)Lehmann, Isele, Jakob, Jentzsch, Kontokostas, Mendes, Hellmann, Morsey, Van~Kleef, Auer, and Bizer]{lehmann14}
Jens Lehmann, Robert Isele, Max Jakob, Anja Jentzsch, Dimitris Kontokostas, Pablo Mendes, Sebastian Hellmann, Mohamed Morsey, Patrick Van~Kleef, Sören Auer, and Christian Bizer.
\newblock Dbpedia - a large-scale, multilingual knowledge base extracted from wikipedia.
\newblock \emph{Semantic Web Journal}, 6, 01 2014.
\newblock \doi{10.3233/SW-140134}.

\bibitem[Li and Yang(2018)]{li18}
Yang Li and Tao Yang.
\newblock \emph{Word Embedding for Understanding Natural Language: A Survey}, pages 83--104.
\newblock Springer International Publishing, Cham, 2018.
\newblock ISBN 978-3-319-53817-4.
\newblock \doi{10.1007/978-3-319-53817-4_4}.
\newblock URL \url{https://doi.org/10.1007/978-3-319-53817-4_4}.

\bibitem[Liang et~al.(2023)Liang, Gonen, Mao, Hou, Goyal, Ghazvininejad, Zettlemoyer, and Khabsa]{liang23}
Davis Liang, Hila Gonen, Yuning Mao, Rui Hou, Naman Goyal, Marjan Ghazvininejad, Luke Zettlemoyer, and Madian Khabsa.
\newblock Xlm-v: Overcoming the vocabulary bottleneck in multilingual masked language models, 2023.
\newblock URL \url{https://arxiv.org/abs/2301.10472}.

\bibitem[List(2012)]{list12}
Johann-Mattis List.
\newblock Lexstat: automatic detection of cognates in multilingual wordlists.
\newblock pages 117--125, 04 2012.

\bibitem[List(2014)]{list14}
Johann-Mattis List.
\newblock Investigating the impact of sample size on cognate detection.
\newblock \emph{Journal of Language Relationship}, 11\penalty0 (1):\penalty0 91--102, 2014.
\newblock \doi{doi:10.31826/jlr-2014-110111}.
\newblock URL \url{https://doi.org/10.31826/jlr-2014-110111}.

\bibitem[List et~al.(2018)List, Greenhill, Anderson, Mayer, Tresoldi, and Forkel]{list18-clics}
Johann-Mattis List, Simon~J. Greenhill, Cormac Anderson, Thomas Mayer, Tiago Tresoldi, and Robert Forkel.
\newblock Clics2: An improved database of cross-linguistic colexifications assembling lexical data with the help of cross-linguistic data formats.
\newblock \emph{Linguistic Typology}, 22\penalty0 (2):\penalty0 277--306, 2018.
\newblock \doi{doi:10.1515/lingty-2018-0010}.
\newblock URL \url{https://doi.org/10.1515/lingty-2018-0010}.

\bibitem[List et~al.(2022)List, Forkel, Greenhill, Rzymski, Englisch, and Gray]{list22}
Johann-Mattis List, Robert Forkel, Simon Greenhill, Christoph Rzymski, Johannes Englisch, and Russell Gray.
\newblock Lexibank, a public repository of standardized wordlists with computed phonological and lexical features.
\newblock \emph{Scientific Data}, 9:\penalty0 316, 06 2022.
\newblock \doi{10.1038/s41597-022-01432-0}.

\bibitem[List and Forkel(2024)]{list24}
Mattis List and Robert Forkel.
\newblock Lingpy. a python library for historical linguistics, 2024.
\newblock URL \url{https://lingpy.org}.

\bibitem[Ma et~al.(2024)Ma, ImaniGooghari, Ye, Pei, Asgari, and Schütze]{ma24}
Chunlan Ma, Ayyoob ImaniGooghari, Haotian Ye, Renhao Pei, Ehsaneddin Asgari, and Hinrich Schütze.
\newblock Taxi1500: A multilingual dataset for text classification in 1500 languages, 2024.

\bibitem[Mayer and Cysouw(2014)]{mayer14}
Thomas Mayer and Michael Cysouw.
\newblock Creating a massively parallel {B}ible corpus.
\newblock In Nicoletta Calzolari, Khalid Choukri, Thierry Declerck, Hrafn Loftsson, Bente Maegaard, Joseph Mariani, Asuncion Moreno, Jan Odijk, and Stelios Piperidis, editors, \emph{Proceedings of the Ninth International Conference on Language Resources and Evaluation ({LREC}'14)}, pages 3158--3163, Reykjavik, Iceland, May 2014. European Language Resources Association (ELRA).
\newblock URL \url{http://www.lrec-conf.org/proceedings/lrec2014/pdf/220_Paper.pdf}.

\bibitem[Miller(1995)]{miller95}
George~A. Miller.
\newblock Wordnet: A lexical database for english.
\newblock \emph{Commun. ACM}, 38:\penalty0 39--41, 1995.
\newblock URL \url{https://api.semanticscholar.org/CorpusID:1671874}.

\bibitem[Mortensen et~al.(2018)Mortensen, Dalmia, and Littell]{mortensen18}
David~R. Mortensen, Siddharth Dalmia, and Patrick Littell.
\newblock Epitran: Precision {G2P} for many languages.
\newblock In Nicoletta Calzolari~(Conference chair), Khalid Choukri, Christopher Cieri, Thierry Declerck, Sara Goggi, Koiti Hasida, Hitoshi Isahara, Bente Maegaard, Joseph Mariani, H\'el\`ene Mazo, Asuncion Moreno, Jan Odijk, Stelios Piperidis, and Takenobu Tokunaga, editors, \emph{Proceedings of the Eleventh International Conference on Language Resources and Evaluation (LREC 2018)}, Paris, France, May 2018. European Language Resources Association (ELRA).
\newblock ISBN 979-10-95546-00-9.

\bibitem[Navigli and Ponzetto(2012)]{navigli12}
Roberto Navigli and Simone Ponzetto.
\newblock Babelnet: The automatic construction, evaluation and application of a wide-coverage multilingual semantic network.
\newblock \emph{Artificial Intelligence}, 193:\penalty0 217–250, 12 2012.
\newblock \doi{10.1016/j.artint.2012.07.001}.

\bibitem[Navigli et~al.(2021)Navigli, Bevilacqua, Conia, Montagnini, and Cecconi]{navigli21}
Roberto Navigli, Michele Bevilacqua, Simone Conia, Dario Montagnini, and Francesco Cecconi.
\newblock Ten years of babelnet: A survey.
\newblock pages 4559--4567, 08 2021.
\newblock \doi{10.24963/ijcai.2021/620}.

\bibitem[Nesterenko et~al.(2024)Nesterenko, Blassel, Veber, Boussau, and Jacob]{nesterenko24}
Luca Nesterenko, Luc Blassel, Philippe Veber, Bastien Boussau, and Laurent Jacob.
\newblock {Phyloformer: Fast, accurate and versatile phylogenetic reconstruction with deep neural networks}.
\newblock working paper or preprint, November 2024.
\newblock URL \url{https://hal.science/hal-04795764}.

\bibitem[{NLLB Team} et~al.(2022){NLLB Team}, Costa-jussà, Cross, Çelebi, Elbayad, Heafield, Heffernan, Kalbassi, Lam, Licht, Maillard, Sun, Wang, Wenzek, Youngblood, Akula, Barrault, Mejia-Gonzalez, Hansanti, Hoffman, Jarrett, Sadagopan, Rowe, Spruit, Tran, Andrews, Ayan, Bhosale, Edunov, Fan, Gao, Goswami, Guzmán, Koehn, Mourachko, Ropers, Saleem, Schwenk, and Wang]{nllb22}
{NLLB Team}, Marta~R. Costa-jussà, James Cross, Onur Çelebi, Maha Elbayad, Kenneth Heafield, Kevin Heffernan, Elahe Kalbassi, Janice Lam, Daniel Licht, Jean Maillard, Anna Sun, Skyler Wang, Guillaume Wenzek, Al~Youngblood, Bapi Akula, Loic Barrault, Gabriel Mejia-Gonzalez, Prangthip Hansanti, John Hoffman, Semarley Jarrett, Kaushik~Ram Sadagopan, Dirk Rowe, Shannon Spruit, Chau Tran, Pierre Andrews, Necip~Fazil Ayan, Shruti Bhosale, Sergey Edunov, Angela Fan, Cynthia Gao, Vedanuj Goswami, Francisco Guzmán, Philipp Koehn, Alexandre Mourachko, Christophe Ropers, Safiyyah Saleem, Holger Schwenk, and Jeff Wang.
\newblock No language left behind: Scaling human-centered machine translation.
\newblock 2022.

\bibitem[Pan et~al.(2017)Pan, Zhang, May, Nothman, Knight, and Ji]{pan17}
Xiaoman Pan, Boliang Zhang, Jonathan May, Joel Nothman, Kevin Knight, and Heng Ji.
\newblock Cross-lingual name tagging and linking for 282 languages.
\newblock In Regina Barzilay and Min-Yen Kan, editors, \emph{Proceedings of the 55th Annual Meeting of the Association for Computational Linguistics (Volume 1: Long Papers)}, pages 1946--1958, Vancouver, Canada, July 2017. Association for Computational Linguistics.
\newblock \doi{10.18653/v1/P17-1178}.
\newblock URL \url{https://aclanthology.org/P17-1178}.

\bibitem[Pompei et~al.(2011)Pompei, Loreto, and Tria]{pompei11}
Simone Pompei, Vittorio Loreto, and Francesca Tria.
\newblock On the accuracy of language trees.
\newblock \emph{PloS one}, 6:\penalty0 e20109, 06 2011.
\newblock \doi{10.1371/journal.pone.0020109}.

\bibitem[Raganato et~al.(2017)Raganato, Camacho-Collados, and Navigli]{raganato17}
Alessandro Raganato, Jose Camacho-Collados, and Roberto Navigli.
\newblock Word sense disambiguation: A unified evaluation framework and empirical comparison.
\newblock In Mirella Lapata, Phil Blunsom, and Alexander Koller, editors, \emph{Proceedings of the 15th Conference of the {E}uropean Chapter of the Association for Computational Linguistics: Volume 1, Long Papers}, pages 99--110, Valencia, Spain, April 2017. Association for Computational Linguistics.
\newblock URL \url{https://aclanthology.org/E17-1010}.

\bibitem[Rama and List(2019)]{rama19}
Taraka Rama and Johann-Mattis List.
\newblock An automated framework for fast cognate detection and {B}ayesian phylogenetic inference in computational historical linguistics.
\newblock In Anna Korhonen, David Traum, and Llu{\'\i}s M{\`a}rquez, editors, \emph{Proceedings of the 57th Annual Meeting of the Association for Computational Linguistics}, pages 6225--6235, Florence, Italy, July 2019. Association for Computational Linguistics.
\newblock \doi{10.18653/v1/P19-1627}.
\newblock URL \url{https://aclanthology.org/P19-1627}.

\bibitem[Sagart et~al.(2019)Sagart, Jacques, Lai, Ryder, Thouzeau, Greenhill, and List]{sagart19}
Laurent Sagart, Guillaume Jacques, Yunfan Lai, Robin Ryder, Valentin Thouzeau, Simon~J. Greenhill, and Johann-Mattis List.
\newblock Dated language phylogenies shed light on the ancestry of sino-tibetan.
\newblock \emph{Proceedings of the National Academy of Science of the United States of America}, 116:\penalty0 10317--10322, 2019.
\newblock \doi{https://doi.org/10.1073/pnas.1817972116}.
\newblock URL \url{https://www.pnas.org/content/early/2019/04/30/1817972116}.

\bibitem[Sharma et~al.(2021)Sharma, Dhawan, and Pailla]{sharma21}
Rahul Sharma, Kunal Dhawan, and Balakrishna Pailla.
\newblock Phonetic word embeddings, 2021.
\newblock URL \url{https://arxiv.org/abs/2109.14796}.

\bibitem[Speer et~al.(2017)Speer, Chin, and Havasi]{speer17}
Robyn Speer, Joshua Chin, and Catherine Havasi.
\newblock Conceptnet 5.5: An open multilingual graph of general knowledge, 2017.
\newblock URL \url{http://aaai.org/ocs/index.php/AAAI/AAAI17/paper/view/14972}.

\bibitem[Strube and Ponzetto(2006)]{strube06}
Michael Strube and Simone~Paolo Ponzetto.
\newblock Wikirelate! computing semantic relatedness using wikipedia.
\newblock In \emph{Proceedings of the 21st National Conference on Artificial Intelligence - Volume 2}, AAAI'06, page 1419–1424. AAAI Press, 2006.
\newblock ISBN 9781577352815.

\bibitem[Suchanek et~al.(2024)Suchanek, Alam, Bonald, Chen, Paris, and Soria]{suchanek24}
Fabian Suchanek, Mehwish Alam, Thomas Bonald, Lihu Chen, Pierre-Henri Paris, and Jules Soria.
\newblock Yago 4.5: A large and clean knowledge base with a rich taxonomy, 2024.

\bibitem[Swadesh(1955)]{swadesh55}
Morris Swadesh.
\newblock Towards greater accuracy in lexicostatistic dating.
\newblock \emph{International Journal of American Linguistics}, 21\penalty0 (2):\penalty0 121--137, 1955.

\bibitem[Sérasset(2012)]{serasset12}
Gilles Sérasset.
\newblock Dbnary: Wiktionary as a lmf based multilingual rdf network.
\newblock 05 2012.

\bibitem[Trost et~al.(2023)Trost, Haag, Höhler, Jacob, Stamatakis, and Boussau]{trost23}
Johanna Trost, Julia Haag, Dimitri Höhler, Laurent Jacob, Alexandros Stamatakis, and Bastien Boussau.
\newblock {Simulations of Sequence Evolution: How (Un)realistic They Are and Why}.
\newblock \emph{Molecular Biology and Evolution}, 41\penalty0 (1):\penalty0 msad277, 12 2023.
\newblock ISSN 1537-1719.
\newblock \doi{10.1093/molbev/msad277}.
\newblock URL \url{https://doi.org/10.1093/molbev/msad277}.

\bibitem[Vossen(1998)]{vossen98}
Piek Vossen, editor.
\newblock \emph{EuroWordNet: a multilingual database with lexical semantic networks for European Languages}.
\newblock Kluwer, 1998.

\bibitem[Zhang et~al.(2017)Zhang, Li, and Wang]{zhang17}
Li~Zhang, Jun Li, and Chao Wang.
\newblock Automatic synonym extraction using word2vec and spectral clustering.
\newblock In \emph{2017 36th Chinese Control Conference (CCC)}, pages 5629--5632, 2017.
\newblock \doi{10.23919/ChiCC.2017.8028251}.

\bibitem[Zouhar et~al.(2024)Zouhar, Chang, Cui, Carlson, Robinson, Sachan, and Mortensen]{zouhar24}
Vilém Zouhar, Kalvin Chang, Chenxuan Cui, Nathaniel Carlson, Nathaniel Robinson, Mrinmaya Sachan, and David Mortensen.
\newblock Pwesuite: Phonetic word embeddings and tasks they facilitate, 2024.

\end{thebibliography}
\end{document}